\pdfoutput=1

\documentclass[11pt]{article}

\usepackage[final]{acl}

\usepackage{times}
\usepackage{latexsym}
\usepackage[table]{xcolor}
\usepackage{enumitem}
\usepackage[T1]{fontenc}

\usepackage[utf8]{inputenc}

\usepackage{microtype}

\usepackage{inconsolata}
\usepackage{microtype}
\usepackage{graphicx}
\usepackage{booktabs}
\usepackage{longtable}
\usepackage{booktabs} 
\usepackage{enumitem}
\usepackage{pgf-pie}
\usepackage{tikz} 
\usepackage{array}
\usepackage{tcolorbox}
\usepackage{multirow}
\usepackage{setspace} 
\usepackage{pgf-pie}
\usepackage{url} 
\usepackage{float} 
\usepackage{hyperref}  
\usepackage[linesnumbered,ruled,vlined]{algorithm2e}
\usepackage{amsmath}
\usepackage{amssymb}
\usepackage{mathtools}
\usepackage{amsthm}

\usepackage{tcolorbox}
\usepackage{subcaption}  
\usepackage{stfloats}

\usepackage{enumitem}
\setlist{nosep, 
         topsep=0pt, 
         partopsep=0pt,
         parsep=0pt,
         itemsep=0pt} 
\title{TurnBack: A Geospatial Route Cognition Benchmark for Large Language Models through Reverse Route}

\author{\textbf{Hongyi Luo}$^{1,2}$ \quad \textbf{Qing Cheng}$^{1,2}$ \quad \textbf{Daniel Matos}$^{1}$ \quad \textbf{Hari Krishna Gadi}$^{1}$ \\
\textbf{Yanfeng Zhang}$^{1}$ \quad \textbf{Lu Liu}$^{1}$ \quad \textbf{Yongliang Wang}$^{1}$ \quad \textbf{Niclas Zeller}$^{3}$ \\
\textbf{Daniel Cremers}$^{2,4}$ \quad \textbf{Liqiu Meng}$^{2}$ \\
$^{1}$Huawei Riemann Lab\thanks{This research was led by the Huawei Riemann Lab.}
\quad $^{2}$Technische Universität München
\quad $^{3}$Hochschule Karlsruhe
\quad $^{4}$MCML \\
\texttt{\{hongyi.luo, qing.cheng1, daniel.matos, hari.krishna.gadi1\}@huawei.com} \\
\texttt{\{zhangyanfeng8, luliu1, wangyongliang775\}@huawei.com} \\
\texttt{niclas.zeller@h-ka.de, \{cremers, liqiu.meng\}@tum.de}
}

\begin{document}
\maketitle
\begin{abstract}
Humans can interpret geospatial information through natural language, while the geospatial cognition capabilities of Large Language Models (LLMs) remain underexplored. Prior research in this domain has been constrained by non-quantifiable metrics, limited evaluation datasets and unclear research hierarchies. Therefore, we propose a large-scale benchmark and conduct a comprehensive evaluation of the geospatial route cognition of LLMs. We create a large-scale evaluation dataset comprised of 36000 routes from 12 metropolises worldwide. Then, we introduce PathBuilder, a novel tool for converting natural language instructions into navigation routes, and vice versa, bridging the gap between geospatial information and natural language. Finally, we propose a new evaluation framework and metrics to rigorously assess 11 state-of-the-art (SOTA) LLMs on the task of route reversal. The benchmark reveals that LLMs exhibit limitation to reverse routes: most reverse routes neither return to the starting point nor are similar to the optimal route. Additionally, LLMs face challenges such as low robustness in route generation and high confidence for their incorrect answers. Code\ \&\ Data available here: \href{https://github.com/bghjmn32/EMNLP2025_Turnback}{TurnBack.}
 
\end{abstract}

\section{Introduction}

\label{submission}

Geospatial cognition is crucial for enabling LLMs to perform advanced route navigation and urban planning, such as "Take me home, pass by any supermarket, and find a mailbox within 500m of it." Humans naturally complete such tasks by relying on their innate geospatial cognitive abilities, which enable them to reason about geospatial relationships based solely on linguistic cues. Equipping LLMs with sophisticated geospatial cognition can significantly enable real-life applications. Recent research indicates that LLMs are able to encode geospatial knowledge \cite{openai2024gpt4technicalreport,fu2024scenellmextendinglanguagemodel,hong20233d,liu2025st}. However, despite these domain-specific advances, a unified hierarchical framework for evaluating geospatial cognition in LLMs remains absent, making the realization of advanced geospatial reasoning an open research challenge.

The geospatial cognitive hierarchy proposed by \citep{werner1997spatial}, widely accepted across geoinformatics and cognitive science for decades \cite{yang2025evaluating}, provides an ideal structure for such evaluation. This framework delineates three hierarchical levels as Figure \ref{fig:landmark_Route_Survey}:

\begin{figure}[H]
  \centering
  \includegraphics[width=0.9\columnwidth]{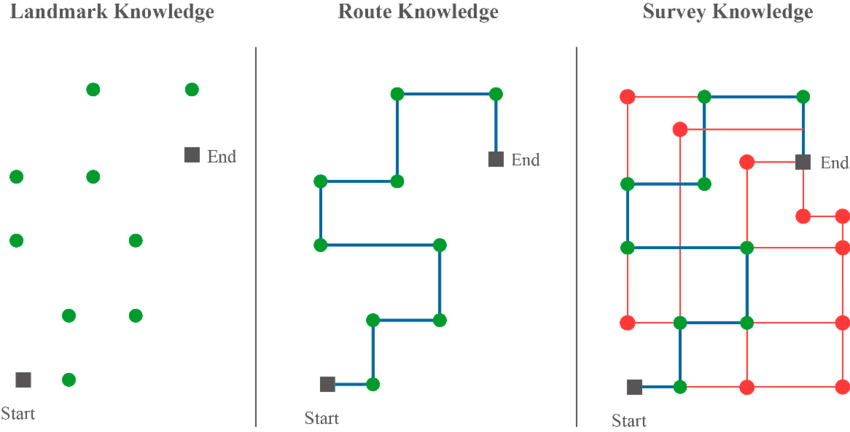}
  \caption{Geospatial cognition hierarchy from Points $\rightarrow$ Routes $\rightarrow$ Networks \cite{quesnot2014measure}. 
  \textbf{Landmark level}—knowledge of static landmarks (e.g.\ a building’s address); 
  \textbf{Route level}—understanding of connections between landmarks, including the length of a route; 
  \textbf{Survey level}—comprehensive geospatial knowledge that enables identification of any landmark and planning routes between them (note that route planning belongs to the \emph{survey}, not the \emph{route}).}
  \label{fig:landmark_Route_Survey}
\end{figure}

Current research disproportionately focuses on Landmark-level geospatial cognition, likely because such knowledge is easier to textualise, yet the same body of work indicates that LLMs face significant cognitive challenges with Route-level knowledge \citep{momennejad2023evaluatingcognitivemapsplanning, gisexam, feng_et_al:LIPIcs.GIScience.2023.28}. Through our experiments and literature review, we found that current LLMs lack map-like survey knowledge. However, they can interpret simpler route information to a certain degree. Therefore, we set up this benchmark with a focus on the Route Knowledge.

\begin{figure*}[htb]  
  \centering

  \begin{subfigure}[t]{0.48\textwidth}
    \centering
    \includegraphics[width=\linewidth]{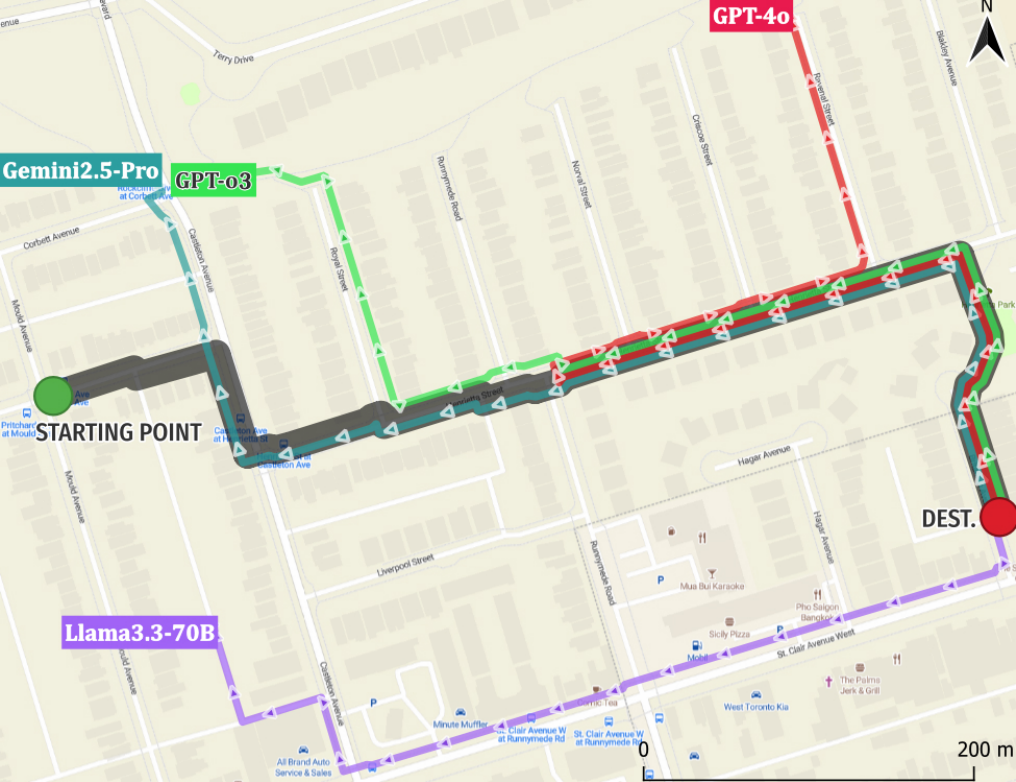}
    \caption{Route reversal of LLMs}
    \label{fig:routereversal:LLMs}
  \end{subfigure}\hfill
  \begin{subfigure}[t]{0.48\textwidth}
    \centering
    \includegraphics[width=\linewidth]{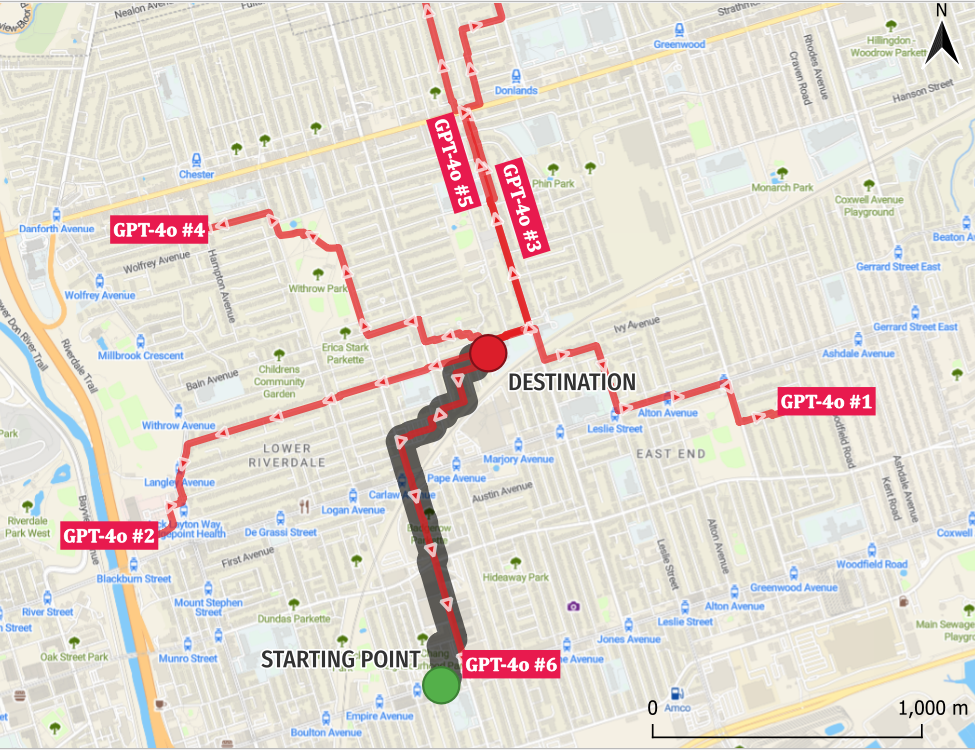}
    \caption{Robustness of route reversal by GPT-4o}
    \label{fig:routereversal:robust}
  \end{subfigure}

  \caption{
    \textbf{a)} In an ``Easy'' route reversal task, LLMs exhibited low accuracy and consistency. None of the reverse routes return to the starting point. Gemini achieved the highest similarity score of 73.4, whereas Llama attained a similarity score of 22.6. 
    \textbf{b)} After six iterations of the same ``Hard'' route reversal task, GPT-4o exhibited significant dissimilarity. The robustness score for the reversed routes is 23.6.
    (\textcopyright~\href{https://www.maptiler.com/copyright/}{MapTiler} \& \href{https://www.openstreetmap.org/copyright}{OpenStreetMap})
  }
  \label{fig:routerevsesal} 
\end{figure*}

Route Knowledge task emphasize sequence-based navigation, landmark-dependent instructions, and procedural paths that do not require global geospatial cognition. Among them, route reversal has been recognized as a representative task for geospatial cognition evaluation within cognitive science and control systems research \cite{furgale2010visual,karimpur2016finding}. It plays a crucial role in evaluating human geospatial cognition \cite{allison2017route} and has contributed to the development of animal-inspired navigation algorithms for robots \citep{kumar2018visualmemoryrobustpath}. The route reversal task involves two distinct conceptual subtasks: (1) spatially determining the endpoint relative to the starting location from a given forward route, and (2) effectively navigating back to the original starting point \citep{behaviourcong}. Unlike landmark knowledge, which can be easily conveyed through textual description, route reversal inherently demands geospatial reasoning without requiring the comprehensive survey knowledge typically associated with route planning. Consequently, route reversal currently presents the most compelling and targeted scenario for evaluating route-level geospatial cognition for LLMs.

\begin{figure*}[htbp]
    \centering
    \includegraphics[width = 1\textwidth]{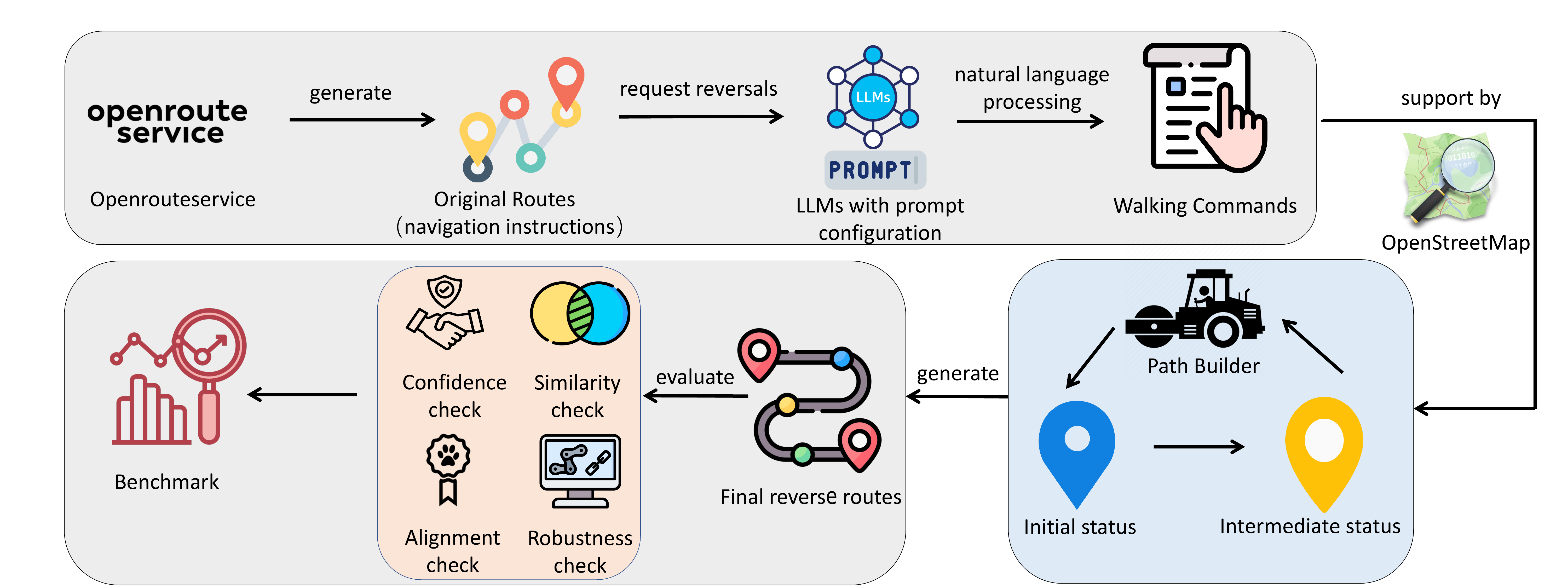}
    \caption{A three-stage workflow: (1)~\textbf{Data Generation}: A route engine generates random routes and collects navigation instructions of route reversal via a prompt-based LLM; (2)~\textbf{Routes Construction}: A Path-Builder constructs reversed routes with geometric support from the OpenStreetMap; (3)~\textbf{Evaluation \& Benchmarking}: Following a multimetric evaluation, a comparison with the geospatial route cognition capabilities of SOTA LLMs is performed.. 
    (\textcopyright~Icons by \href{https://www.flaticon.com}{Flaticon})}
    \label{fig:longworkflows}
\end{figure*}

In this paper, we introduce TurnBack, a benchmark explicitly designed to evaluate the route-level geospatial cognition of LLMs on the route reversal task. To achieve this, we develop an algorithm leveraging OpenStreetMap and OpenRouteService \cite{openrouteservice} to generate extensive, realistic route datasets complete with navigational instructions. LLMs are then tasked with generating reversed navigational instructions, to guide the construction of reversed routes using our novel tool, PathBuilder (PB). Furthermore, we propose a comprehensive evaluation framework to systematically measure the performance of SOTA LLMs on this task and analyse the disorders. The complete workflow for our proposed approach is illustrated in Figure \ref{fig:longworkflows}.
Our contributions are as follows:
\begin{itemize}[nosep]
\item \textbf{TurnBack Benchmark}: We release the first large-scale route-reversal dataset including 36\,000 pedestrian routes across 12 global cities at three difficulty levels. We also provide a comprehensive evaluation schema to reveal the performance of LLMs. It offers a reproducible probe of route-level geospatial cognition in LLMs.
\item \textbf{PathBuilder}: a novel language-to-geometry converter. It bridges the gap between the formal language of route geometry and the natural language processing capabilities of LLMs.
\item \textbf{Comprehensive Disorders Study}: We benchmark nine SOTA LLMs and expose four recurring geospatial cognition disorders. We point out that LLM currently suffers from architectural weaknesses in geospatial cognition.
\end{itemize}

\section{Related Work}
\textbf{LLMs' Geospatial Cognition}: LLMs perform well on \textbf{Landmark} cognition tasks like answering geographic questions \cite{bhandari2023large}, but struggle with \textbf{Survey} cognition tasks such as route planning and navigation \cite{mansourian2024chatgeoai, yan2024georeasoner, gupta2024geode}. While some work shows LLMs can internalize spatial representations like latitude and longitude \cite{gurnee2023language}, their cognition is still in an early transition from Landmark to Route knowledge, facing significant challenges.

\textbf{Route Reversal Benchmark}: Existing benchmarks for LLM geospatial cognition have three common limitations. First, \textbf{Landmark Overfocus}: Many studies rely on repetitive landmark questions that play to the LLMs' language strengths \cite{manvi2024geollmextractinggeospatialknowledge, mooney2023towards}. Second, \textbf{Premature Survey Inquiry}: Some works test survey knowledge without first ensuring the model has robust route knowledge \cite{ding2024mango}. Third, \textbf{Vision Confusion}: The use of vision makes it hard to attribute performance to either perception or internal reasoning \cite{feng2024citybench}. In contrast, route reversal is a long-standing metric in geospatial research that predates LLMs \cite{behaviourcong, allison2017route, donald2002self, coutrot2022entropy}. It effectively isolates route-based cognition from both vision and landmark knowledge. See Appendix \ref{Related_Work} for more details.

\section{Benchmark}
\subsection{TurnBack Dataset}

\label{sec:datageneration}

The dataset in this paper covers all continents (except Antarctica), with two representative metropolitan cities in each continent: Toronto, Denver, Mexico City, São Paulo, London, Munich, Tokyo, Singapore, Sydney, Auckland, Cairo and Cape Town. 3,000 routes were extracted from each city and equally divided into 3 difficulty levels, resulting in 36,000 routes in total, with our proposed algorithm~\ref{alg:path_builder}. Routes selected for this study range between 500 and 2,500 meters, suitable for pedestrian navigation with OpenStreetMap and OpenRouteService \cite{openrouteservice}. 

\subsubsection{Data Generation}

The dataset creation involves five steps: (1) generating starting points $S_i$ following a Gaussian distribution within a city, with a particular latitude and longitude as its center; (2) selecting endpoints $D_j$ randomly within a specified radius $(r_{min}, r_{max})$ from each $S_i$; (3) computing routes and extracting navigation instructions for each valid pair $(S_i, D_j)$; (4) standardizing instructions through natural language processing; and (5) compiling formatted instructions with corresponding route geometry data. Theoretically, this dataset can be scaled infinitely as long as computational resources allow.

\subsubsection{Data Split}
Human geospatial cognition performance can be influenced by different urban road network patterns \cite{coutrot2022entropy}. In order to reveal the fine-grained performance with regard to the road pattern, we need to classify routes into different difficulty levels. Thus, we propose a simple method to measure the complexity of a route using two fundamental metrics: length, and number of turns, as shown in Eq. \ref{eq:difficulty}. 

\begin{table}[t]
    \centering
    \scalebox{0.9}{
        \begin{tabular}{lccc}
            \toprule
            Difficulty & Samples & Avg. Length & Avg. Turns \\
            \midrule
            Easy      & $12000$   & $925$       & $4.6$      \\
            Medium    & $12000$   & $1598$      & $7.8$      \\
            Hard      & $12000$   & $2032$      & $13.2$     \\
            \bottomrule
        \end{tabular}
    }
    \caption{Dataset characteristics across difficulty levels. Each level is defined by equally dividing the 0-100\% range, with 5\% buffer zones at transitions. }
    \label{tab:dataset_stats}
\end{table}

\begin{equation}
\label{eq:difficulty}
C = \frac{d_{\text{max}} - \frac{n_t}{l} }{d_{\text{max}} - d_{\text{min}}} \times 100 
\end{equation}
where \( n_t \) denotes the number of turns in the route and \( l \) is the route length in meters. The parameters \( d_{\text{min}} \) and \( d_{\text{max}} \) are the minimum and maximum ratios of \( \frac{n_t}{l} \) within the dataset. The complexity \( C \) is normalized to a range from 0 to 100.

We partitioned the dataset into three difficulty levels (easy, medium, and hard) by equally distributing them according to the complexity. This systematic approach ensures clear difficulty demarcation. As shown in Table~\ref{tab:dataset_stats}, the dataset characteristics vary significantly across difficulty levels. In addition, interesting geographic heterogeneity does exist across cities, see Table \ref{tab:route_city}.

\begin{algorithm}[t]
\small 
\caption{PathBuilder: Navigation Instruction to Geometric Path}
\label{alg:path_builder}
\KwIn{Navigation instructions $I$, OpenStreetMap data}
\KwOut{Geometric path $P$}

\BlankLine 

\textbf{Phase 1:} Parse Instructions\;
Convert natural language instructions to command sequence $C$\;
Initialize starting position $S$ and direction $\theta$ (North: $0^\circ$)\;
$Q \leftarrow \emptyset$ \tcp*{Init coord. queue} 

\BlankLine

\textbf{Phase 2:} Process Commands\;
\ForEach{command $c_i \in C$}{
    \uIf{$c_i$ is turn command}{
        Update $\theta$ based on turn angle\;
    }
    \uElseIf{$c_i$ is move command}{
        Update position $S$ based on $\theta$ and distance\;
        $Q \leftarrow Q \cup \{S\}$ \tcp*{Store new pos.} 
    }
}

\BlankLine

\textbf{Phase 3:} Generate Path\;
Connect points in $Q$ using routing engine\;

\BlankLine

\KwRet{$P$}
\end{algorithm}

\subsubsection{PathBuilder}

This section presents a novel tool that facilitates the construction of path based on natural language instructions or vice versa. Current routing engines, such as Google Maps or OpenRouteService, exhibit varied navigation instruction styles but generally adhere to common formatting principles. Frequent use of verbs like ``turn'', ``go'', ``keep'', and ``continue'' enables these instructions to be translated into sample commands for path construction. Geometrically, a route is represented by a sequence of connected points, with each navigation command guiding the selection of subsequent points based on specific rules. Thus, replicating a route involves reversing the original translation from geometry to natural language. As outlined in Algorithm~\ref{alg:path_builder}, the path-building process comprises three main phases.

We evaluated the PathBuilder's performance using similarity scores across diverse urban environments, specifically in Tokyo, Munich, and Toronto, see Appendix \ref{PBevaluation}. The results show that it is a powerful tool for generating geometry from navigation instruction.

\subsection{Evaluation Metrics}
The metrics employed comprise two aspects: (1) geometric performance, and (2) LLMs' generation performance. They evaluate the route level knowledge of LLMs in terms of both the quality of the returned geometry as well as their thinking process.

\subsubsection{Geometric Performance} 
There is an important assumption: since our original route was generated using a route-planning engine, the optimal reverse route should be backtracked. As Figure \ref{fig:datageneration:routereversal} shows, 
any reverse route that differs from the original one and does not return to the start point is considered a failure to some extent.      
Thus, we have two ground truth references: the starting point and the optimal reverse route.

\textbf{Return Rate}: percentage of reverse routes that return to the start point (tolerances of up to 20 meters are allowed). 

\textbf{Similarity Score}\label{similaritycalculation}: We define the similarity measure \( \text{sim}(x_i, x_j) \) as a weighted sum of multiple geographical and mathematical metrics, where \( x_i \) and \( x_j \) denote two routes being compared. The metrics include Length Ratio (LR), Hausdorff Distance (HD), Fréchet Distance (FD), Edit Distance (ED), Jaccard Index (JI), Angle (A), and Sum of Coordinates Offsets (SCO). The similarity score is given by:
\begin{equation}
\text{Sim}(x_i, x_j) = w_{\text{LR}} \cdot \text{LR} + w_{\text{HD}} \cdot \text{HD} + \dots + w_{\text{SCO}} \cdot \text{SCO}
\label{eq:similarity}
\end{equation}
where \( w_k \) represent their respective weight.

In our evaluation, a similarity score above 80 indicates strong resemblance, scores over 90 suggest near equivalence, while values below 50 denote dissimilarity, and scores under 30 indicate completely distinct routes. Further details on the metrics can be found in Table~\ref{tab:Similarity_calculation_metrix}.

\subsubsection{Generation Performance} f

\textbf{Robustness}: We evaluate the consistency of responses generated by LLMs for the same routing task using a robustness score. We do so by computing the standard deviation \(\sigma\) of pairwise similarities among responses:

\begin{equation}
\sigma = \sqrt{\frac{2}{N(N-1)} \sum_{i=1}^{N-1} \sum_{j=i+1}^{N} \left(\text{sim}(x_i, x_j) - S\right)^2},
\end{equation}
where \(S\) is the average similarity across all unique response pairs. Finally, \(\sigma\) is normalized using min-max normalization over the full dataset to yield a robustness score \(R\in[0,100]\), where higher values indicate greater consistency and robustness.

\textbf{Confidence}: Inspired by \citet{xu2024earthflatbecauseinvestigating}, we adapt their method 
to compute the confidence level of LLMs in the generated instructions.

\begin{equation} \label{eq:confidenceappendix}
\mathit{Confidence}_{set} = \frac{1}{N} \sum_{i=1}^{N} \mathit{DirectionProb}_i
\end{equation}
where \(N\) is the total number of direction instructions, and \(\mathit{DirectionProb}_i\) is the token probability of the direction word in the \(i\)-th instruction. This probability can be calculated from the log probabilities record of LLMs. For example, if the token “north” has an 80\% likelihood among candidate tokens, we interpret the LLM as being 80\% confident in “turning north.” The detailed confidence calculation is provided in Appendix~\ref{confidencecheck}.

\textbf{Misalignment}: As an important metric of LLM research, we decide to use misalignment to measure the performance of LLMs to output valid instructions. In our task, this metric is not about correctness but represents the LLMs' capacity for geospatial reasoning. Even after prompt engineering, they may still produce invalid navigation instructions or generate broken routes.

\subsection{Prompt Design}
Given that prompt engineering significantly affects the model's focus and returned content format \cite{he2024doespromptformattingimpact}, we use two types of prompts: the guide prompt, and the instruction prompt.

\textbf{The guide prompt, shown in Figure \ref{fig:initial_prompt}}, is meant to let the LLMs focus on the geographic information of the current experiment location. It attempts to ensure that the LLM does not use text-based semantic inversion methods because we found that LLMs tend to invert direction words to answer such as “turn left and walk 500 meters southeast” to “turn right and walk 500 meters northwest”. 

\textbf{The instruction prompt, shown in Figure \ref{fig:instructions_prompt}},is intended to align the return format of the LLM with the navigation instructions that the PB can execute. For each navigation instruction, the coordinates and nearby landmarks (if any) are given. This is to encourage LLMs to think with the geospatial information they have been trained with.

\subsection{Landmark Preliminary Experiment}
As an enhancement to the benchmark, we constructed 200 manually curated question sets for preliminary evaluation, illustrating that current LLMs exhibit geospatial cognition situated between advanced landmark knowledge and route knowledge. LLMs perform well on knowledge with abundant textual support but struggle with lesser-known landmarks. Notably, performance degrades sharply when tasked with directional reasoning or coordinate system calculation. Due to space limitations, details are provided in Appendix \ref{landmark_geo}.

\section{Route Reversal Benchmark}
We use \textbf{Reverse Route} denotes the answers returned by LLMs, \textbf{Original Route} to represent the sample route sent to LLMs for route reversal. In this paper we use Original Route as the ground truth for route reversal evaluation. Because all routes are optimized by the route planning engine in the generation process, the optimal reverse route is the original route backwards.

\subsection{Target LLMs}
We tested eleven SOTA LLMs:GPT-4o, GPT-o1, GPT-O3, Gemini1.5-pro, Gemini2.5-Pro, Llama3.3-70B, Deepseek(R1), Claude 3.5,Claude 3.7 and Grok. All open-source models were set up according to the HuggingFace tutorial and temperature was set 0. Temperature is proved not a determinant of performance, see Appendix \ref{temperature_scetion},\ref{experiment_setup}.

\begin{table*}[t!] 
  \centering
  \caption{Benchmark of LLMs on route reversal task (geometric performance).}
  \small
  \setlength{\tabcolsep}{2pt}

  \begin{tabular}{@{} l c c c c c c @{}}
    \toprule
    \textbf{Model}
      & \textbf{Return Rate}
      & \textbf{Similarity}
      & \textbf{Deviation Angle}
      & \textbf{Hausdorff Distance}
      & \textbf{Length Ratio}
      & \textbf{Jaccard Index} \\
      & (\%) $\uparrow$
      & $\uparrow$
      & $\downarrow$ ($^\circ$)
      & $\downarrow$ (m)
      & ($\leftrightarrow$1)
      & $\uparrow$ \\
    \midrule
    \rowcolor{green!10}
    GPT-4o        & 6.34  & 41.06 (0.26) & 32.18 (0.23) & 169.47 (1.07) & 0.88 (2.4e-3) & 0.36 (1.6e-3) \\
    GPT-o1        & 9.47    & 48.13 (0.24)      & 29.47 (0.22)       & 142.65 (0.89)       & 0.71 (2.9e-3)        & 0.43 (2.4e-3) \\
    \rowcolor{green!10}
    GPT-o3        & 12.25 & 53.99 (0.24) & \underline{23.22} (0.36) & 123 (0.62) & \underline{0.71} (2.9e-3) & 0.43 (2.4e-3) \\
    Gemini1.5-Pro & 11.93    & 61.71 (0.19)      & 36.63 (0.25)       & 136.23 (0.85)   & 1.12 (3.2e-3)        & 0.50 (3.1e-2)\\
    \rowcolor{green!10}
    Gemini2.5-Pro & \textbf{14.46} & \textbf{67.26} (0.15) & 26.73 (0.32) & 113.23 (0.55) & 1.01 (2.2e-1) & \textbf{0.53} (2.7e-3) \\
    Llama3.3-70B  & \underline{4.06}  & 42.78 (0.27) & \textbf{53.67} (0.35) & \underline{189.32} (1.19) & 0.79 (3.4e-3) & 0.37 (2.8e-3) \\
    \rowcolor{green!10}
    Deepseek      & 7.63  & \underline{40.01} (0.16) & 31.23 (0.23) & 152.80 (0.94) & 0.93 (4.2e-3) & \underline{0.34} (2.6e-3) \\
    Deepseek R1   & 9.42  & 48.15 (0.19) & 30.35 (0.20) & 131.21 (0.78) & 1.09 (3.1e-3) & 0.41 (2.3e-3) \\
    \rowcolor{green!10}
    Claude 3.5    & 7.33  & 40.62 (0.17) & 36.83 (0.25) & 158.39 (1.03) & 0.93 (2.2e-3) & 0.35 (1.5e-3) \\
    Claude 3.7    & 9.05  & 49.50 (0.20) & 33.15 (0.20) & \textbf{128.56} (0.81) & \textbf{1.17} (5.5e-3) & 0.44 (3.8e-3) \\
    \rowcolor{green!10}
    Grok          & 6.72  & 40.79 (0.29) & 33.42 (0.23) & 183.85 (1.16) & 0.77 (3.1e-3) & 0.35 (2.7e-3) \\
    \midrule
    \rowcolor{yellow!10}
    GPT-4o        & 2.93  & 36.19 (0.19) & 84.17 (0.57) & 340.28 (2.17) & 1.06 (2.2e-3) & 0.31 (1.2e-3) \\
    GPT-o1        & 3.68    & 43.87 (0.19)      & 67.12 (0.47)       & 289.17 (1.81)       & 1.05 (3.6e-3)        & 0.39 (2.2e-3) \\
    \rowcolor{yellow!10}
    GPT-o3        & 5.16  & 47.16 (0.19) & 63.85 (0.46) & 283.90 (1.76) & 1.05 (3.0e-3) & 0.41 (2.2e-3) \\
    Gemini1.5-Pro & 5.12    & 47.34 (0.14)      & 68.20 (0.49)       & 287.90 (1.84)       & 1.08 (3.1e-3)        & 0.40 (2.1e-3) \\
    \rowcolor{yellow!10}
    Gemini2.5-Pro & \textbf{7.63} & \textbf{51.33} (0.14) & 67.42 (0.47) & 279.76 (1.72) & \textbf{1.22} (3.1e-3) & \textbf{0.42} (2.1e-3) \\
    Llama3.3-70B  & \underline{1.82} & 39.29 (0.13) & \textbf{104.25} (0.72) & \textbf{392.23} (2.47) & \underline{1.00} (3.6e-3) & 0.34 (2.0e-3) \\
    \rowcolor{yellow!10}
    Deepseek      & 1.79  & 38.23 (0.12) & 66.95 (0.48) & 311.49 (1.93) & 1.13 (6.2e-3) & \underline{0.32} (3.4e-3) \\
    Deepseek R1   & 4.09  & 44.32 (0.14) & 67.73 (0.52) & 278.14 (1.73) & 1.10 (5.1e-3) & 0.38 (3.5e-3) \\
    \rowcolor{yellow!10}
    Claude 3.5    & 3.18  & 39.16 (0.13) & 78.92 (0.53) & 372.55 (2.33) & 1.19 (7.2e-3) & 0.33 (3.9e-3) \\
    Claude 3.7    & 3.92  & 44.16 (0.14) & \underline{56.73} (0.42) & \underline{274.62} (1.70) & 1.07 (4.4e-3) & 0.38 (2.8e-3) \\
    \rowcolor{yellow!10}
    Grok          & 2.96  & \underline{37.16} (0.23) & 99.87 (0.67) & 365.41 (2.23) & 1.07 (4.1e-3) & 0.32 (2.4e-3) \\
    \midrule
    \rowcolor{red!10}
    GPT-4o        & 0.21  & 24.13 (0.15) & 132.15 (0.92) & 789.23 (5.02) & 1.19 (3.3e-3) & 0.23 (1.2e-3) \\
    GPT-o1        & 0.26  & 29.91 (0.16) & 98.20 (0.75)  & 705.40 (4.60) & 1.16 (5.0e-3) & 0.26 (2.0e-3) \\
    \rowcolor{red!10}
    GPT-o3        & 0.57  & 33.81 (0.16) & 91.87 (0.70) & 682.57 (4.29) & \textbf{1.31} (5.4e-3) & 0.28 (2.1e-3) \\
    Gemini1.5-Pro & 0.37  & 29.56 (0.10) & 90.18 (0.67) & 670.38 (4.20) & 1.02 (4.0e-3) & 0.27 (2.3e-3) \\
    \rowcolor{red!10}
    Gemini2.5-Pro & 0.57  & 34.01 (0.10) & 96.15(0.61) & 550.38 (3.25) & 0.91 (4.1e-3) & 0.26 (2.4e-1) \\
    Llama3.3-70B  & 0.49  & 24.87 (0.10) & \textbf{156.93} (1.07) & \textbf{810.45} (5.25) & 1.25 (7.5e-3) & 0.22 (2.5e-3) \\
    \rowcolor{red!10}
    Deepseek      & \textbf{0.87} & 26.86 (0.10) & 102.82 (0.78) & 733.61 (4.57) & 0.98 (3.2e-3) & 0.25 (1.6e-3) \\
    Deepseek R1   & 0.18 & \textbf{30.10} (0.10) & 120.93 (0.87) & \underline{623.15} (3.86) & 1.05 (4.1e-3) & \textbf{0.28} (2.2e-3) \\
    \rowcolor{red!10}
    Claude 3.5    & 0.83  & 26.10 (0.10) & 120.78 (0.85) & 747.82 (4.78) & 1.18 (6.4e-3) & 0.24 (2.4e-3) \\
    Claude 3.7    & 0.61  & 28.18 (0.09) & \underline{86.47} (0.73) & 672.41 (4.18) & 1.09 (5.2e-3) & 0.26 (2.4e-3) \\
    \rowcolor{red!10}
    Grok          & \underline{0.14}  & \underline{23.98} (0.21) & 144.28 (1.08) & 747.56 (4.78) & \underline{0.85} (3.1e-3) & \underline{0.21} (1.5e-3) \\
    \bottomrule
  \end{tabular}

  \vspace{1ex}
  \footnotesize{Note: The values in parentheses are standard errors. The highest and lowest are shown in bold and underlined, respectively. Green, yellow, and red indicate easy, medium, and difficult, respectively (same as Table \ref{tab:by_difficulty_side_by_side}). The Hausdorff distance can be seen as the distance between two geometries. See Appendix \ref{similaritymetrics} for more details.} 
  \label{tab:llm_performance}
\end{table*}

\subsection{Results and Findings}

\textbf{Results:} The main results of SOTA LLMs on the proposed benchmark are summarized in Table \ref{tab:llm_performance} and  \ref{tab:by_difficulty_side_by_side}. As depicted in Figure \ref{fig:model_performance}, the similarity scores are derived from the normalized cumulative distributions of similarity. Each route was tested 6 times.

Our analysis indicates that route reversals generated by LLMs generally exhibit low return rates, low similarity, and low robustness despite high confidence scores.

\textbf{Findings I:} Current SOTA LLMs struggle considerably with the route reversal task, as exemplified in Figure \ref{fig:routereversal:LLMs}. Even in the easy dataset, involving routes typically shorter than 1 km with 3–5 turns, LLMs only achieve reture rates between 4.0\% and 11.9\%. For routes of medium complexity, most models fail entirely. This observed difficulty with basic route reversal tasks underscores a critical limitation in current LLMs' geospatial route cognition. Consequently, among routes that fail to return, the similarity between them ground truth is also quite low.

\textbf{Findings II:} LLMs suffer from \textbf{inconsistency} in geospatial route cognition. Routes generated by identical prompts of the same LLM exhibit notable inconsistencies, resulting in low robustness across all tested models. Gemini-2.5-pro achieved the highest robustness score at 69.15, while Llama-3.3-70B had the lowest at only 38.58. These results clearly indicate that current LLMs display significant stochasticity in their geospatial route cognition. Figure \ref{fig:routereversal:robust} shows an example of the inconsistency generation of GPT-4o. Theoretically, if LLMs have a certain level of knowledge about the question, there should be coherence among multiple responses, even if they are incorrect. However, we observe that LLMs' multiple responses are not convergent, which means that the route generated by their multiple responses have very little similarity to each other.

\textbf{Findings III:} Obvious \textbf{misalignment} demonstrated by LLMs is their notable deficiency in identifying essential information for constructing valid routes, frequently leading to the omission of critical initial absolute directional guidance. An absolute direction (e.g., east, northwest, or $100^\circ$) at the starting point is required to establish a viable route. Our experiments reveal that this inability to provide an actionable first step accounts for 4--16\% of route generation failures. Furthermore, some LLM responses degenerate into complete failures, offering overly generic or non-navigational advice, exemplified by suggestions like, "From your starting location, simply walk back to your destination." Notably, unlike common alignment challenges, where iterative prompting can rectify errors, repeated emphasis on the necessity of an initial absolute direction yields only marginal improvements in this context. This persistent deficiency suggests a more fundamental limitation: LLMs may lack an internalized spatial framework crucial for geolocation-based reasoning, particularly in grasping the importance of "self-localization" and "sequential spatial linkage" for effective route construction.

\textbf{Findings IV:} Despite general inadequacies, more advanced LLMs demonstrate clear advantages in geospatial tasks. Gemini2.5-pro, previously recognized for its superior spatial cognitive capabilities \cite{yang2024thinkingspacemultimodallarge}, notably outperforms other models across some metrics. The contrast between GPT-o3 and GPT-4o further illustrates this point: GPT-o3, considered more advanced in reasoning, surpasses GPT-4o by 5.9\% in return rate and 13\% in similarity on the easy dataset, alongside better robustness. However, it is interesting to note that chain-of-thought could make the model performance drop on difficult datasets, for example, Deepseek vs. R1. These observations confirm both the ongoing progress of LLMs in geospatial cognition and the discriminative effectiveness of our benchmark.

\begin{table*}[htbp]
  \centering
  \caption{Benchmark of LLMs on route reversal task (disorder performance).}
  \renewcommand{\arraystretch}{0.8}  
  \setlength{\tabcolsep}{2pt}       
  \scriptsize

  \noindent\makebox[\textwidth][c]{%
    \begin{minipage}[t]{0.35\textwidth}
      \centering
      \textbf{Easy}\\[0.7pt]
      \begin{tabular}{c@{\hspace{1pt}}ccc}
        \toprule
        Model & Robustness & Confidence & Misalignment \\
        \midrule
        \rowcolor{green!15}
        GPT-4o        & 43.74 & 92.50 & \textbf{14.57} \\
        \rowcolor{white}
        GPT-o1        & 51.32    & 96.26    & 8.03   \\
        \rowcolor{green!15}
        GPT-o3        & 67.49 & \textbf{96.51} & 5.16  \\
        \rowcolor{white}
        Gemini1.5-Pro & 45.79    & 93.97    & 7.52   \\
        \rowcolor{green!15}
        Gemini2.5-Pro & \textbf{69.15} & 91.23 & 3.51  \\
        \rowcolor{white}
        Llama3.3-70B  & \underline{38.58} & \underline{86.52} & 12.53 \\
        \rowcolor{green!15}
        Deepseek      & 47.32 & 90.04 & 12.58 \\
        \rowcolor{white}
        Deepseek R1   & 52.41 & 93.18 & 9.53  \\
        \rowcolor{green!15}
        Claude 3.5    & 53.12 & N/A   & 9.14  \\
        \rowcolor{white}
        Claude 3.7    & 56.13 & N/A   & \underline{4.58}  \\
        \rowcolor{green!15}
        Grok          & 44.16 & N/A   & 10.07 \\
        \bottomrule
      \end{tabular}
    \end{minipage}\hspace{4pt}%
    \begin{minipage}[t]{0.35\textwidth}
      \centering
      \textbf{Medium}\\[2pt]
      \begin{tabular}{c@{\hspace{1pt}}ccc}
        \toprule
        Model & Robustness & Confidence & Misalignment \\
        \midrule
        \rowcolor{yellow!15}
        GPT-4o        & 36.24 & 91.76 & 13.04 \\
        \rowcolor{white}
        GPT-o1        & 44.9    & 92.57    & 8.08   \\
        \rowcolor{yellow!15}
        GPT-o3        & 52.16 & \textbf{93.61} & 7.83  \\
        \rowcolor{white}
        Gemini1.5-Pro & 37.52    & 87.46    & 6.19   \\
        \rowcolor{yellow!15}
        Gemini2.5-Pro & \textbf{56.15} & \textbf{90.15} & 6.94  \\
        \rowcolor{white}
        Llama3.3-70B  & \underline{27.40} & \underline{85.63} & 11.06 \\
        \rowcolor{yellow!15}
        Deepseek      & 38.12 & 91.23 & \textbf{13.12} \\
        \rowcolor{white}
        Deepseek R1   & 36.58 & 92.63 & 8.08  \\
        \rowcolor{yellow!15}
        Claude 3.5    & 43.59 & N/A   & 9.56  \\
        \rowcolor{white}
        Claude 3.7    & 46.39 & N/A   & \underline{5.06}  \\
        \rowcolor{yellow!15}
        Grok          & 40.29 & N/A   & 11.59 \\
        \bottomrule
      \end{tabular}
    \end{minipage}\hspace{4pt}%
    \begin{minipage}[t]{0.33\textwidth}
      \centering
      \textbf{Hard}\\[2pt]
      \begin{tabular}{c@{\hspace{1pt}}ccc}
        \toprule
        Model & Robustness & Confidence & Misalignment \\
        \midrule
        \rowcolor{red!15}
        GPT-4o        & \underline{19.74} & 89.47 & 16.12 \\
        \rowcolor{white}
        GPT-o1        & 38.85    & 94.25    & 7.15   \\
        \rowcolor{red!15}
        GPT-o3        & \textbf{42.16} & \textbf{89.72} & 6.75  \\
        \rowcolor{white}
        Gemini1.5-Pro & 37.52    & 87.46    & 6.19   \\
        \rowcolor{red!15}
        Gemini2.5-Pro & 40.09 & \underline{88.51} & 7.02  \\
        \rowcolor{white}
        Llama3.3-70B  & 13.60 & 92.15 & \textbf{16.21} \\
        \rowcolor{red!15}
        Deepseek      & 21.54 & 90.25 & 14.57 \\
        \rowcolor{white}
        Deepseek R1   & 20.04 & 93.16 & 9.49  \\
        \rowcolor{red!15}
        Claude 3.5    & 32.13 & N/A   & 9.03  \\
        \rowcolor{white}
        Claude 3.7    & 39.57 & N/A   & \underline{5.12}  \\
        \rowcolor{red!15}
        Grok          & 37.51 & N/A   & 10.54 \\
        \bottomrule
      \end{tabular}
    \end{minipage}%
  }

  \label{tab:by_difficulty_side_by_side}
\end{table*}

\section{Geospatial Cognition Disorder Study}   
The failure pattern led us to identify two disorders unique to geospatial cognition: disorientation and superficiality. The distribution of the above disorders is shown in Table \ref{tab:geospatial_cognition_disorders}. Due to space limits, we select four typical models to analysis.

\begin{table}[H]
    \centering
    \caption{Geospatial Cognition Disorders for Route Reversal (a route can suffer multiple disorders).}
    \scriptsize 
    \setlength{\tabcolsep}{25pt} %
    \begin{tabular}{lc}
        \toprule
        \textbf{Disorder}    & \textbf{Proportion} \\
        \midrule
        Inconsistency        & 47\%                \\
        Superficiality          & 21\%                \\
        Misalignment         & 12\%                 \\
        Disorientation       & 63\%                \\
        \bottomrule
    \end{tabular}
    \label{tab:geospatial_cognition_disorders}
\end{table}

\textbf{Disorientation} not only means that LLMs get lost in road networks, but also that they cannot identify current location from the starting point. According to the low return rate, most routes generated end up far from the ground truth. This indicates that LLMs have disorders in constructing the road network in their latent space. Moreover, we found that even within a single instruction, LLMs might behave in a confusing manner when questions are asked about the current location. In an example from GPT-4o, after “Turn west and continue for 100 m,” followed by “Turn right and go straight for 300 m,” GPT-4o claimed that the endpoint lay \emph{southwest} of the origin, an impossible result. Repeating this probe on 300 easy routes and checking the answers manually, we found that GPT-4o identified the final direction correctly in only about 50 \% of the cases (Table~\ref{tab:disorientation_gpt4o}), even when the task only require an approximate direction.

\begin{table}[H]
\centering
\caption{Disorientation of GPT4o for 300 Samples}
\label{tab:disorientation_gpt4o}
\scriptsize  
\setlength{\tabcolsep}{25pt} 
\begin{tabular}{lc}
\toprule
\textbf{Difficulty Level} & \textbf{Disorientation (\%)} \\
\midrule
Easy (100)    & 32\% \\
Medium (100)  & 56\% \\
Hard (100)    & 85\% \\
\bottomrule
\end{tabular}
\end{table}

\textbf{Superficiality} is a tendency to employ task-agnostic heuristics or exploit statistical patterns in the training data, rather than engaging in the problem-specific reasoning required by the prompt. In the context of route reversal tasks, genuine problem solving necessitates geospatial reasoning, i.e. understanding relative positions. However, LLMs often exhibit superficiality by resorting to semantic inversion---a simple reversal of directional terms ("north" to "south", "left" to "right") and step order---without necessarily processing the underlying spatial relationships. Crucially, this semantic inversion heuristic does not reliably guarantee accuracy and represents a shortcut around the intended computational process.

Our experiments reveal a notable correlation between this superficial behavior and the models' confidence: outputs generated via semantic inversion consistently exhibit higher token probabilities compared to outputs from geospatial reasoning. As illustrated in Table \ref{tab:llm_confidence}, we suggest that token probability analysis could serve as an indicator for detecting whether an LLM is employing domain-specific reasoning or relying on superficial heuristics.

\begin{table}[htbp]
\centering
\caption{Average Confidence Differences of LLMs in Semantic Reversal (superficial approach) and Normal Geospatial Reasoning in 200 samples}
\label{tab:llm_confidence}
\resizebox{\columnwidth}{!}{%
\begin{tabular}{lcccc}
\toprule
\textbf{Reasoning Mode} & \textbf{GPT4o} & \textbf{Llama3.3-70B} & \textbf{Gemini2.5 Pro} & \textbf{GPTo3} \\
\midrule
Semantic Reversal       & 98             & 99                   & 98                    & 99            \\
Geospatial Thinking         & 92             & 91                   & 93                    & 90            \\
\bottomrule
\end{tabular}%
}
\end{table}

\subsection{Discussion}
\label{sec:discussion}
Our analysis suggests that poor LLM performance on route-reversal tasks is attributable to a fundamental architectural deficit in geospatial cogntion, rather than to easily remedied factors like prompt engineering. This claim is supported by our empirical results: low return rates and similarity, high variance across trials, and systematically inflated confidence for invalid solutions. Moreover, the marginal improvement from prompt refinements suggests the bottlenecks are systemic and lie deeper than surface-level instruction following.

\textbf{Representation bottleneck.}
Sub-word tokenizers divide coordinates such as52.5167 into ``52'', ``.'', ``516'', ``7'', mapping them to nearly orthogonal vectors. This process breaks the relationship between a number and its representation, violating the principle of \emph{metric continuity}. Consequently, a small numerical change becomes a large, unrelated jump in the embedding space. This lack of numerical sensitivity creates cascading errors in operations that depend on metric continuity, such as heading updates or cumulative offsets, amplifying small inaccuracies into large route deviations. Studies confirm that LLMs' grasp of ordered magnitude is weak and heavily dependent on the tokenizer \cite{yang2024number,sessler2024benchmarking}. This explains why models generate semantically plausible but metrically divergent paths.

\textbf{Objective misalignment.}
The next-token cross-entropy objective is agnostic to numerical proximity. For example, predicting 52.5168° incurs the same loss as 90.00° when the true value is 52.5167°. Because the loss function is defined over a discrete vocabulary, the optimization gradient cannot signal the \emph{magnitude} of numerical error. The model is thus incentivized to match tokens for syntactic fluency, not to minimize the metric distance between its predicted coordinates and the ground truth. This explains the observed gap between high token probabilities (confidence) and low geometric fidelity. This finding aligns with prior work, indicating that LLMs degrade to random performance in tasks requiring sub-degree accuracy \cite{kazemi2023geomverse}.

\textbf{Data sparsity.}
High-precision coordinates and long-tail place names are exceedingly rare in web-scale text, causing their token frequencies to approach zero. Unlike specialized models that discretize space into learnable grid cells (e.g., H3) at the cost of precision \cite{schestakov2024trajectory}, LLMs lack a principled spatial hashing mechanism. Consequently, they default to coarse linguistic heuristics, such as inverting ``left/right'' and reordering steps. This strategy achieves textual plausibility but fails to reconstruct the fine-grained geometric path. This outcome is consistent with the performance drop on harder routes where small metric errors accumulate.

\textbf{Failure of Geometric Compositionality.}
The self-attention mechanism, which computes a content-weighted average, is fundamentally a semantic aggregator, not a geometric transformer. It lacks the inductive bias for affine transformations like rotation, translation, and scaling that are foundational to path integration \cite{dziri2023faithfatelimitstransformers}. This architectural deficit directly causes the empirical failures observed in our benchmark, such as \emph{disorientation} (failure to maintain a global heading) and initial \emph{misalignment}. The model resorts to shallow symbolic flips, a phenomenon termed ``linearized subgraph matching'' \cite{dziri2023faithfatelimitstransformers}, such as swapping ``left/right'' without re-grounding the trajectory. This indicates a reliance on surface patterns over a latent spatial scaffold.

Taken together, these factors offer a coherent explanation for the observed failures. Crucially, they clarify why stronger prompting yields only limited gains. Prompts can steer output format, but they cannot supply the numeric continuity, metric-aware objectives, dense spatial data, or geodesic operators that are missing from the model's architecture. Addressing these gaps is critical for enabling reliable route-level geospatial cognition.

\section{Possible Mitigation}
\label{sec:mitigation}

\begin{figure}[H]
  \centering
  \includegraphics[width=0.85\columnwidth,keepaspectratio]{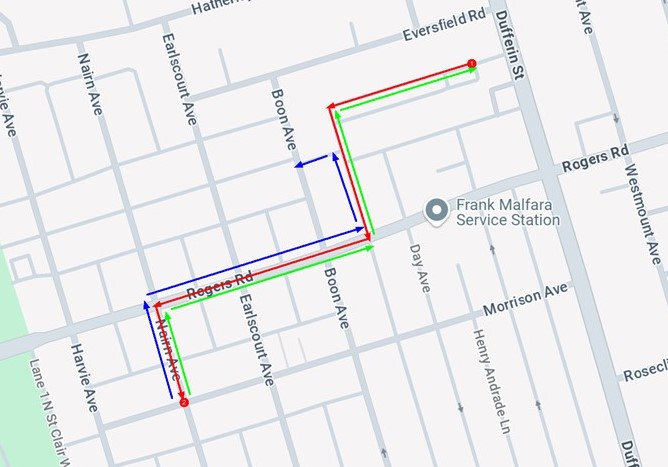}
  \caption{Original route (red), reverse path without image prompt (blue), and with the vector-map prompt (green) on an easy sample in Toronto.}
  \label{fig:mitgration-toronto}
\end{figure}

Full architectural changes (numeric-aware tokenizers, metric-aware losses, spatial operators) are costly. We therefore test a low-cost, inference-time aid without modifying our language-first benchmark. We attach a \emph{vector map} to the original prompt. The map is rendered from OSM-style vector tiles, shows streets and salient POIs only. Reversal rules and output format remain unchanged. We report two settings: \textbf{Original (language-only)} for route cognition without visual confounds, and \textbf{Assisted (language + vector map)} as a minimal cue for inference. On 200 \emph{easy} routes in Toronto (GPT-4o), the return rate rises from 6.4\% to 43.7\%, and similarity from 41.06 to 73.08 (Fig.~\ref{fig:mitgration-toronto}, Table~\ref{tab:gpt4o_image_prompt}).

\begin{table}[t]
    \centering
    \caption{Effect of adding a vector map to the prompt (GPT-4o, 200 easy routes, Toronto).}
    \label{tab:gpt4o_image_prompt}
    \resizebox{0.90\columnwidth}{!}{%
    \begin{tabular}{lcc}
        \toprule
        \textbf{Condition}           & \textbf{Return Rate (\%)} & \textbf{Similarity} \\
        \midrule
        Original         & 6.4                       & 41.06               \\
        Assisted            & 43.7                      & 73.08               \\
        \bottomrule
    \end{tabular}}%
\end{table}

The gain shows that lightweight visual context reduces gross reversal failures. It does not establish the mechanism: improvements may come from spatial abstraction or from visual anchoring that bypasses textual reasoning. We use vector maps (topology and labels, but no imagery) to limit spurious cues while a full disentanglement is left for future analysis. Making maps the \emph{primary} input would mix visual perception with route cognition and weaken attribution. We insist on a language-first solution because it isolates route reasoning and aligns with the landmark–route–survey hierarchy. There is evidence that route knowledge can be acquired without vision \cite{tinti2006visual} in cognition science research. So while we admit that the assisted setting can improves utility from engineering perspective, it does not address the architectural issues in Sec.~\ref{sec:discussion}.

\section{Conclusion}
We introduced a benchmark \textit{TurnBack} for evaluation on geospatial route cognition. It reveals that SOTA LLMs are still far from reliable geospatial reasoners. On the \textit{TurnBack}, no model solved the route reversal task properly. Even on "easy" dataset the return rate was below 12\%, and performance collapsed to near zero on harder routes. Reverse routes often failed to reach the start point, and repeated queries produced divergent paths, revealing the absence of a stable internal geospatial representation. Moreover, models expressed unwarranted confidence, underscoring a deep misalignment between natural language generation and true geospatial cognition.

Models that pass the benchmark will prove competent in route-cognition tasks such as road-network knowledge graph and show potential in route planning. By turning these weaknesses into quantifiable analyses, we highlight three research avenues: (i) representations that preserve distance—numeric-aware tokenizers or geospatial cell vocabularies; (ii) objectives that punish metric error rather than pure string mismatch; and (iii) inductive biases that embed geodesic operators or call external map tools. Such advances can potentially enables LLMs achieve reliable geospatial cognition. 

\section{Limitations}
Although our approach offers novel insights into evaluating LLMs' geospatial route cognition, several limitations warrant acknowledgment.

\textbf{Data Constraints}: Our dataset, comprising 36,000 routes across 12 cities on 6 continents, offers substantial coverage. However, it does not encompass the full diversity of global urban road network typologies. Furthermore, other network types, such as those in rural areas or national parks, which could elicit different geospatial reasoning from LLMs, were not included. This exclusion was due to the significant challenges in data acquisition and quality assurance required for a comprehensive Earth-scale exploration within the constraints of this study.
It is pertinent to note that our data generation methodology, leveraging OpenStreetMap (OSM), theoretically permits scaling to OSM's full data. However, this is beyond the practical scope of this research.

\textbf{Model Constraints}: Our empirical evaluation was necessarily confined to a selected set of LLMs due to computational resource limitations. Consequently, the findings may not directly generalize to all existing or future LLMs. The field of LLMs is characterized by rapid advancements, with new models and architectures emerging frequently. While we endeavored to include SOTA and representative models available during our experimental phase (early 2025), it is inevitable that some newer models may not be covered by the time of publication.
Additionally, financial constraints precluded the large-scale evaluation of certain proprietary models, such as GPTo4. Nevertheless, we emphasize that our proposed methodology and its implementation are reproducible with the APIs of a vast majority of LLMs, ensuring the broader applicability of our evaluation framework.

\textbf{PathBuilder Constraints} Due to the topology diversity of the routes and the vague expression of natural language outputs from LLMs, it is challenging to implement a PathBuilder with 100\% reproduction accuracy. Please refer to Appendix \ref{pathbuilderapp} for detailed explanations and examples.

\textbf{Theoretical constraints} Although route reversal is the most representative path knowledge task in our theory, however, we cannot deny that there still exist some other representative tasks such as the interrogation of geometric properties of routes. These tasks are either too trivial or not convincing enough. Ideally all representative tasks from Landmark-Route should be integrated, but such a workload is beyond the scope of this paper.

\subsection{Ethical Considerations}
To the best of our knowledge, we do not have any potential ethical concerns to disclose.




\clearpage  
\appendix

\section{Related Work (Full)}
\label{Related_Work}

\textbf{LLMs' Geospatial Cognition}: 
\textbf{LLMs' Geospatial Cognition}: Recent research has demonstrated that LLMs perform relatively well in \textbf{Landmark} cognition tasks, such as answering geographic questions \cite{bhandari2023large}. Conversely, LLMs generally struggle with \textbf{Survey} cognition tasks, including route planning and urban navigation \cite{mansourian2024chatgeoai, yan2024georeasoner, gupta2024geode}. Although \citet{gurnee2023language} indicate that LLMs can internalize real-world spatial representations (e.g., latitude, longitude, and timestamps), current progress clearly shows that LLM geospatial cognition is transitioning from Landmark knowledge toward Route knowledge, yet significant challenges remain.

\textbf{Route Reversal Benchmark}: A few benchmarks have explored LLM geospatial cognition, yet they typically exhibit three limitations. First, \textbf{Landmark Overfocus}: due to the natural language-friendly nature of LLMs, many studies disproportionately emphasize repetitive landmark knowledge questions \cite{manvi2024geollmextractinggeospatialknowledge, mooney2023towards}. Second, \textbf{Premature Survey Inquiry}: studies such as \cite{ding2024mango} prematurely examine survey knowledge without recognizing that effective route navigation first requires robust route knowledge. Third, \textbf{Vision Confusion}: integrating vision into geospatial tasks complicates the clear attribution of LLM performance to either perceptual reactions or internal geospatial reasoning \cite{feng2024citybench}. In contrast, route reversal tasks have long served as essential metrics in geospatial cognition research, predating the emergence of LLMs \cite{behaviourcong, allison2017route, donald2002self, coutrot2022entropy}, as they strictly involve route-based cognition without reliance on vision or textual knowledge.

\textbf{PathBuilder}: Researchers have made many advances in virtual reality, fintech and text generation on the topic of how to transform formal language between natural language \cite{white2024livebench,yin2024text2vrscene,openai2024gpt4technicalreport}. In the field of route navigation, there are already mature solutions for convert route geometry to natural language \cite{bast2016route,schumann2021generatinglandmarknavigationinstructions}. Since the introduction of LLMs, research in another direction has become particularly important - LLMs are not yet able to understand geometric languages directly. Our work bridges the gap in this area.

\textbf{Geometric similarity} calculation for polylines varies by application \cite{frontiera2008comparison, yang2022geometric, fu2018moment}, and embedding methods can handle scale discrepancies in map data \cite{Hari}.

\textbf{LLMs confidence of generation} has been shown possibly to be artificially interfered with by the prompt \cite{xu2024earthflatbecauseinvestigating}. In \citet{huang2025llm}, \textbf{robustness of LLMs generations} is explored to prove LLMs suffer from uncertain reponses in different task. For \textbf{misalignment}, pervious research have discussed different alignment methods and their effects which is important to the safety of LLMs \cite{shen2023largelanguagemodelalignment,wolf2023fundamental}.

\onecolumn

\section{Data Generation}

\begin{figure*}[htb]
  \centering

  \begin{subfigure}[t]{0.48\textwidth}
    \centering
    \includegraphics[width=\linewidth]{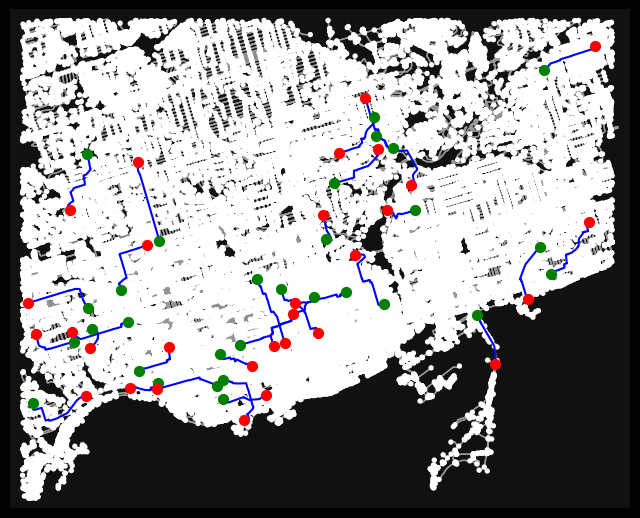}
    \caption{Route generation}
    \label{fig:datageneration:routegen}
  \end{subfigure}\hfill
  \begin{subfigure}[t]{0.48\textwidth}
    \centering
    \includegraphics[width=\linewidth]{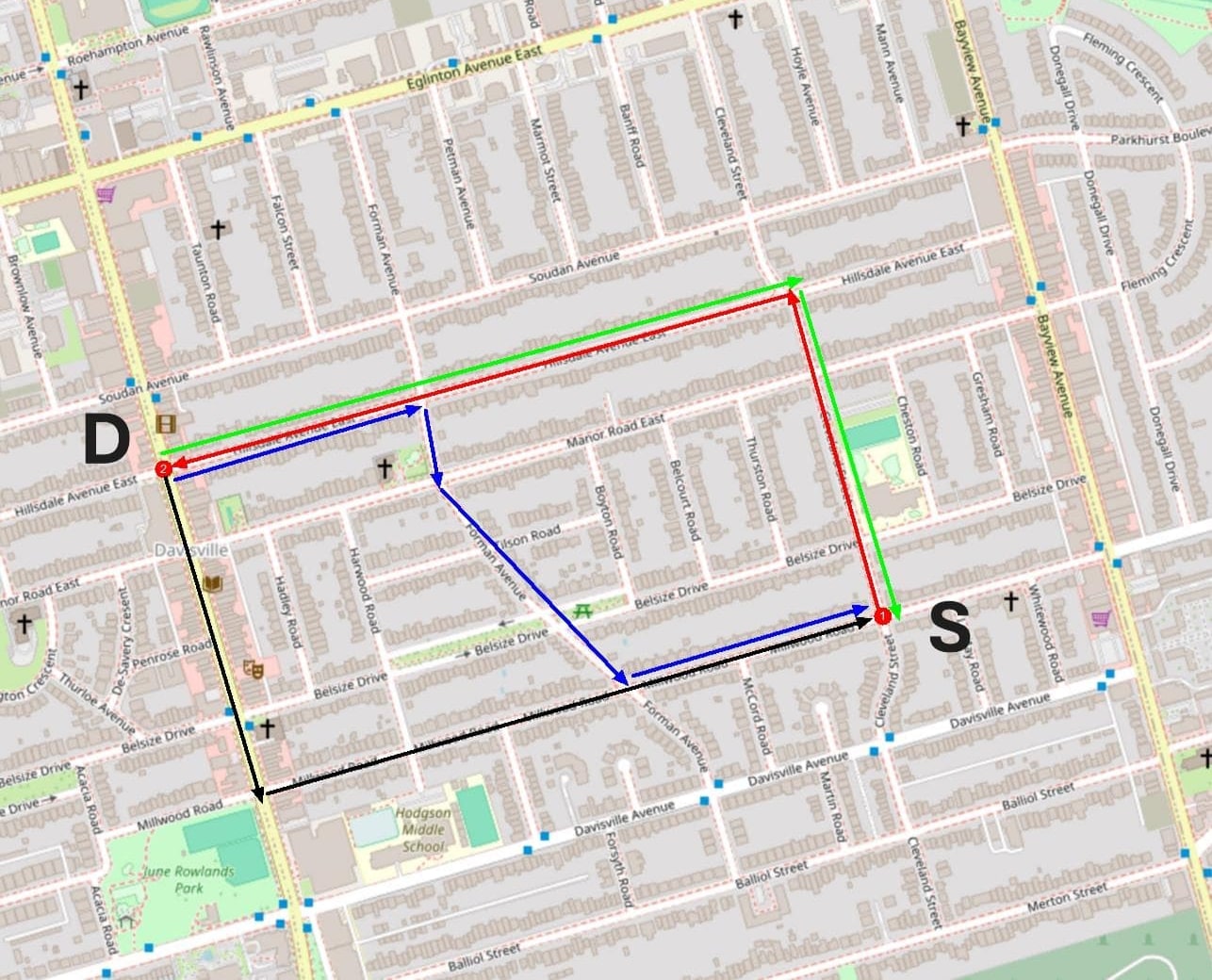}
    \caption{Route reversal}
    \label{fig:datageneration:routereversal}
  \end{subfigure}

  \caption{(Left) 50 routes generated in Toronto between 500 and 2500 meters in length. (Right) The red route between S and D represents the original route; as it is generated by the routing engine, its optimal reverse route is itself. The other returned routes, though valid reversals, differ by varying degrees of similarity.}
  \label{fig:datageneration}
\end{figure*}

\section{Similarity Metrics}
\label{similaritymetrics}
\begin{table*}[!hbt]
    \centering
    \small 
    \renewcommand{\arraystretch}{1.5} 
    \caption{Geographical Measurement Metrics and Their Descriptions}
    \label{tab:Similarity_calculation_metrix}
    \begin{tabular}{|>{\centering\arraybackslash}m{0.35\textwidth}|m{0.55\textwidth}|}
        \hline
        \textbf{\rule{0pt}{3ex} Metrics} & \textbf{\rule{0pt}{3ex} Description} \\
        \hline
        Length Ratio & The ratio of the length of a path or curve to a reference length, commonly used to compare the relative sizes of paths. \\
        \hline
        Hausdorff Distance & The maximum distance between two point sets, defined as the greatest distance from a point in one set to the nearest point in the other set. Used to assess the spatial deviation between two paths. \\
        \hline
        Fréchet Distance & A similarity measure between curves that accounts for both location and orientation, often described as the minimum "travel distance" required for two entities to traverse their respective curves simultaneously. It captures the overall similarity in the shape and traversal order of the paths. \\
        \hline
        Edit Distance & A metric that measures the number of single-point edits (insertions, deletions, or substitutions) required to transform one path's point sequence into another. It evaluates the sequential similarity between two paths. \\
        \hline
        Jaccard Index & A statistic for comparing the similarity and diversity of sample sets, defined as the size of the intersection divided by the size of the union of the sample sets. In path similarity, it measures the overlap between the regions covered by two paths. \\
        \hline
        Angle & The geometric measure of the rotation between two intersecting lines or vectors, typically expressed in degrees or radians. In path similarity, it quantifies the directional difference between corresponding segments of the paths. \\
        \hline
        Sum of Coordinate Offsets & The total sum of the differences between corresponding coordinates (typically in the x and y dimensions) of two geometric objects or paths. This measure is useful for assessing the overall displacement between the objects. \\
        \hline
    \end{tabular}
\end{table*}

\clearpage

\begin{longtable}{llccc}
\caption{Route properties for different levels and different cities.} 
\label{tab:route_city} \\
\toprule
\textbf{Difficulty Level} & \textbf{City} & \textbf{Length (SE) [m]} & \textbf{Turns (SE)} & \textbf{Complexity} \\
\midrule
\endfirsthead

\multicolumn{5}{c}{{\bfseries \tablename\ \thetable{} -- continued from previous page}} \\
\toprule
\textbf{Difficulty Level} & \textbf{City} & \textbf{Length (SE) [m]} & \textbf{Turns (SE)} & \textbf{Complexity} \\
\midrule
\endhead

\midrule \multicolumn{5}{r}{{Continued on next page}} \\
\endfoot

\bottomrule
\endlastfoot

\multirow{12}{*}{Easy} 
& Toronto     & 917.23 (3.63)  & 4.12 (0.04)  & 0.22 \\
& Denver      & 923.91 (3.72)  & 4.26 (0.03)  & 0.21 \\
& Mexico City & 912.03 (3.56)  & 4.50 (0.04)  & 0.24 \\
& São Paulo   & 935.01 (3.78)  & 4.62 (0.03)  & 0.25 \\
& London      & 903.10 (3.44)  & 4.38 (0.04)  & 0.23 \\
& Munich      & 946.53 (3.69)  & 4.51 (0.04)  & 0.22 \\
& Tokyo       & 877.73 (3.47)  & 5.15 (0.04)  & 0.27 \\
& Singapore   & 895.37 (3.66)  & 4.89 (0.04)  & 0.26 \\
& Sydney      & 947.95 (3.56)  & 4.23 (0.03)  & 0.22 \\
& Auckland    & 919.00 (3.75)  & 4.38 (0.03)  & 0.23 \\
& Cairo       & 908.00 (3.59)  & 4.71 (0.04)  & 0.25 \\
& Cape Town   & 915.00 (3.50)  & 4.69 (0.03)  & 0.24 \\

\midrule
\multirow{12}{*}{Medium}
& Toronto     & 1612.63 (5.41) & 7.54 (0.05)  & 0.48 \\
& Denver      & 1623.94 (5.28) & 7.23 (0.05)  & 0.46 \\
& Mexico City & 1598.83 (5.56) & 7.79 (0.06)  & 0.50 \\
& São Paulo   & 1642.94 (5.66) & 7.87 (0.05)  & 0.49 \\
& London      & 1604.04 (5.16) & 7.41 (0.05)  & 0.47 \\
& Munich      & 1681.92 (5.53) & 7.73 (0.06)  & 0.49 \\
& Tokyo       & 1596.48 (5.06) & 8.12 (0.06)  & 0.52 \\
& Singapore   & 1612.33 (5.25) & 7.96 (0.06)  & 0.51 \\
& Sydney      & 1635.27 (5.72) & 7.34 (0.05)  & 0.47 \\
& Auckland    & 1627.00 (5.47) & 7.56 (0.05)  & 0.48 \\
& Cairo       & 1607.00 (5.13) & 7.86 (0.05)  & 0.50 \\
& Cape Town   & 1618.00 (5.41) & 7.68 (0.05)  & 0.49 \\

\midrule
\multirow{12}{*}{Hard}
& Toronto     & 2142.27 (8.03) & 12.77 (0.07) & 0.72 \\
& Denver      & 2122.26 (7.66) & 12.39 (0.06) & 0.70 \\
& Mexico City & 2117.62 (7.88) & 12.91 (0.08) & 0.74 \\
& São Paulo   & 2154.02 (8.06) & 13.26 (0.07) & 0.73 \\
& London      & 2105.27 (7.50) & 12.53 (0.07) & 0.71 \\
& Munich      & 2202.91 (8.00) & 13.18 (0.08) & 0.71 \\
& Tokyo       & 2076.61 (7.34) & 13.58 (0.07) & 0.76 \\
& Singapore   & 2095.83 (7.53) & 13.32 (0.07) & 0.75 \\
& Sydney      & 2136.15 (7.59) & 12.67 (0.07) & 0.72 \\
& Auckland    & 2129.00 (7.69) & 12.68 (0.07) & 0.71 \\
& Cairo       & 2110.00 (7.47) & 13.02 (0.07) & 0.73 \\
& Cape Town   & 2125.00 (7.66) & 12.87 (0.07) & 0.72 \\
\end{longtable}

\twocolumn

\section{PathBuilder discussion and evaluation}
\label{PBevaluation}
In this section, we will evaluate the PathBuilder using the generated dataset. We choose three cities: Tokyo, Munich, and Toronto as the dataset generation locations for this work. This is mainly due to the fact that these cities are spread over three continents, have large populations, and distinct urban scenes.

As shown in Table \ref{tab:PBdataset}, the PathBuilder performs well in all cities. The success rate in these three cities also matches the complexity characteristics of their urban road networks. For example, Tokyo is often considered a city with a very narrow and complex road network \cite{usui2018estimation}. On the other hand, the design of the North American urban road network, represented by Toronto, is generally considered to be more regular \cite{king2020evaluating}. While the PathBuilder is able to handle the majority of cases, there are still some circumstances that prevent it from achieving perfect results. These include theoretically unattainable bottlenecks as well as technical problems that we have not yet solved such as roundabout, but the attained performance is sufficiently good for our task. The details are provided in Appendix \ref{pathbuilderapp}.

\begin{table}[h!]
\centering
\begin{tabular}{|m{1cm}|>{\centering\arraybackslash}m{1.5cm}|>{\centering\arraybackslash}m{1.8cm}|>{\centering\arraybackslash}m{1.8cm}|}
\hline
\textbf{City} & \textbf{Number} & \textbf{Length (m)} & \textbf{Success (\%)} \\
\hline
Toronto & 6000 & 1670 & 96 \\
\hline
Tokyo & 6000 & 1422 & 90 \\
\hline
Munich & 6000 & 1733 & 94 \\
\hline
\end{tabular}
\caption{Length is the average length of all routes; the success rate means it passes the similarity check with a threshold of 85\%.}
\label{tab:PBdataset}
\end{table}

\section{PathBuilder Limitations}
\label{pathbuilderapp}

\begin{figure*}[htbp]
  \centering

  \begin{subfigure}[t]{0.32\textwidth}
    \centering
    \includegraphics[width=\linewidth]{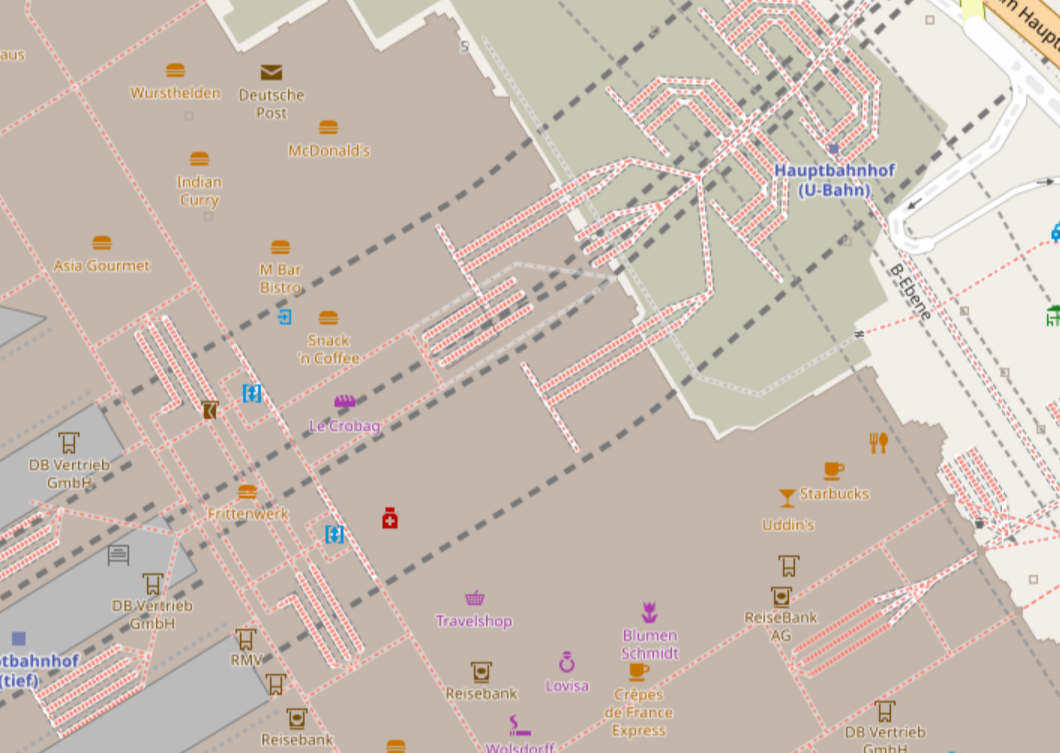}
    \caption{High Complexity Road Network}
    \label{fig:highdensity}
  \end{subfigure}\hfill
  \begin{subfigure}[t]{0.32\textwidth}
    \centering
    \includegraphics[width=\linewidth]{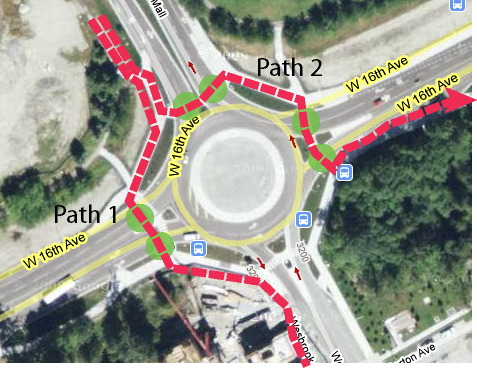}
    \caption{Roundabout}
    \label{fig:roundabout}
  \end{subfigure}\hfill
  \begin{subfigure}[t]{0.32\textwidth}
    \centering
    \includegraphics[width=\linewidth]{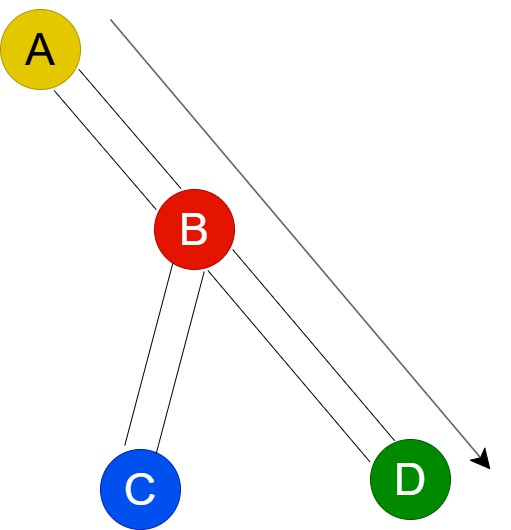}
    \caption{Node of Street}
    \label{fig:street}
  \end{subfigure}

  \caption{Some corner cases remain unsolved or have low accuracy in PathBuilder.}
  \label{fig:subfiguresofPB}
\end{figure*}

\subsection{Information Loss and Semantic Ambiguity}
There have been studies demonstrating the existence of loss of information transfer between (and within) different expression and coding systems \cite{tsvetkov2013information,pang2019towards,cai2018exploring, shannon1993claude}. In our work, geospatial information has actually suffered some loss before entering the PathBuilder: navigation information expressed in natural language does not fully express the original geometric information. In the navigation information from OpenRouteService, the path engine typically describes the magnitude of a turn as “slightly” or “sharply” \cite{openrouteservice}. For example, a turn with an angle between 11° and 44° is called “slightly” in the output instructions. In practice, such information is valid with the added tolerance of human vision in the human eye. However, this does not always apply when transforming from natural language to geometry. In our experiments, we found many incorrect corner cases at boundary values. This type of error can only be optimized by repeated experiments but is very difficult to eradicate. This explains most of the PathBuilder errors.

\subsection{Extremely high density road network}
For these reasons, PB cannot work accurately in a network with high density and accuracy requirements. As shown in Figure \ref{fig:highdensity}, in a train station network in Germany, PB performed poorly despite the fact that the design of the walkway is correct.

\subsection{Mismatch between network nodes and actual streets}

As shown in Figure \ref{fig:street}. Although ABD can be considered as a single street for a pedestrian, point B is treated as a Node in the road network database which means that geometrically BC and BD are two parts. So in reality, if the PB is instructed to “go straight” from A to D through the entire street, it will have to stop at C because it will have to make a “turn” judgment here in the database. Although it is possible to approximate the “next street” implied by the similarity of the angles, but this inevitably weakens the accuracy.

\subsection{Roundabout}

Although rarer, traffic circles are actually scenarios that cannot be resolved in walking mode. The navigation instructions for traffic circles are highly dependent on visual information such as “third exit”, which is very difficult to accomplish from a geospatial perspective. For example, as shown in Fig \ref{fig:roundabout}, current navigation instructions are highly related to vision information such as "Leave roundabout the first exit on your right."

\section{Models Confidence Check}
\label{confidencecheck}
It is well established that humans are more vulnerable to misinformation in complex road networks \cite{KHADEMI201966}. Consequently, it is crucial to gauge the level of confidence LLMs have in their responses to better understand their susceptibility to misinformation.

A previous study by \citet{xu2024earthflatbecauseinvestigating} used token probabilities to estimate how confident LLMs are in their answers, thus offering insight into the models' susceptibility to misinformation. However, because route reversal is not a simple Boolean question, we cannot directly mimic that approach by extracting token probabilities for “yes” or “no.”

In this paper, we adapt their method to accommodate for our task. Specifically, for each navigation instruction generated by LLMs, we compute the confidence by extracting the token probability of the direction word (e.g., “North”) that follows the “turn” action. We then approximate the overall confidence for the entire set of navigation instructions by taking the average of all such direction-word probabilities as Equation \ref{eq:confidenceappendix} shows.

\begin{equation} \label{eq:confidenceappendix}
\mathit{Confidence}_{set} = \frac{1}{N} \sum_{i=1}^{N} \mathit{DirectionProb}_i
\end{equation}

where \(N\) is the total number of direction instructions, and \(\mathit{DirectionProb}_i\) is the token probability of the direction word in the \(i\)-th instruction. This probability can be calculated from the log probabilities record of LLMs.

\section{Experiment setup}
\label{experiment_setup}
The APIs used for all closed-source models are the official APIs for the February 1, 2025 model, with default parameters except for the temperature control, which is zero. For the open-source models, the experimental models were taken from the official version of Hugging Face, and the temperature was also set to 0. The open-source models were set in a training environment of 8 RTX 4090s.

The dataset generation time was about two weeks, mainly due to the limitation on the number of requests from openrouteservice.

\section{Route reversal performance for GPT4o under different temperatures}
\label{temperature_scetion}

To investigate the impact of the temperature parameter on LLMs in the route reversal task, we performed an ablation study using GPT4o at various temperature settings, evaluated on the all city dataset at the easy difficulty level, see Table \ref{tab:temperature_GPT4o}. As expected, lower temperature values produced more deterministic responses, yielding better overall performance. However, even at the lowest temperature (0.0), GPT4o exhibited notable randomness, particularly reflected in robustness and misalignment scores. Thus, we conclude that current LLMs, including GPT4o, lack sufficient spatial cognitive capability to reliably generate consistent route reversal responses. To minimize such randomness and ensure comparability across models, we adopt a temperature setting of 0.0 for all subsequent experiments where applicable. Additionally, as this benchmark aims to fairly assess baseline performance of existing LLMs, no domain adaptation or detailed prompt engineering was employed to artificially boost results.

\begin{table*}[htbp]
    \centering
    \caption{Route reversal performance for GPT4o under different temperature settings. The evaluation was conducted on the all city dataset at easy difficulty level.}
    \label{tab:temperature_GPT4o}
    \resizebox{\textwidth}{!}{%
    \begin{tabular}{cccccc}
        \toprule
        \textbf{Temperature} & \textbf{Return Rate (\%)} & \textbf{Similarity} & \textbf{Robustness} & \textbf{Confidence} & \textbf{Misalignment} \\
        \midrule
        0.0 & 6.70 & 41.12 & 43.50 & 92.60 & 14.20 \\
        0.1 & 6.50 & 41.01 & 42.90 & 92.30 & 14.70 \\
        0.2 & 6.40 & 40.76 & 42.59 & 93.10 & 15.10 \\
        0.3 & 6.20 & 40.30 & 42.31 & 92.77 & 15.40 \\
        0.4 & 6.10 & 40.12 & 41.70 & 93.21 & 16.20 \\
        0.5 & 6.10 & 40.00 & 40.96 & 92.78 & 16.70 \\
        0.6 & 6.10 & 39.96 & 40.42 & 91.87 & 17.10 \\
        0.7 & 5.90 & 39.44 & 40.05 & 93.15 & 17.40 \\
        0.8 & 5.80 & 38.78 & 39.87 & 92.67 & 17.60 \\
        1.0 & 5.50 & 38.54 & 39.23 & 92.31 & 17.90 \\
        \bottomrule
    \end{tabular}%
    }
\end{table*}

\section{Landmark knowledge validation}
\label{landmark_geo}
Although there have been clear indications that LLMs are able to understand landmark knowledge and extract Points of Interest (POIs) \cite{kim2024poi,liu2024semantic}, few datasets have been able to comprehensively examine their abilities. To prove that LLMs adequately grasp Landmark Knowledge, we decide to run a small scale of geospatial landmark knowledge test.

Our dataset contains 100 questions with three dimensions: detailed information about popular Landmarks (40\%), simple information about common Landmarks (30\%), and coordinates transformations related tests (30\%). A large part of our questions is inspired by \citet{yang2025evaluating}.

All responses were manually verified and the results are shown in Table \ref{tab:landmark_model_performance}. The specifics of the questions are available in Appendix \ref{landmark_question}. 

\begin{table}[ht]
    \centering
    \caption{Performance of LLMs in landmark knowledge}
    \label{tab:landmark_model_performance}
    \begin{tabular}{lccc}
        \toprule
        \textbf{LLM} & \textbf{Popular} & \textbf{Common} & \textbf{Geocode} \\
        \midrule
        ChatGPT-4o      & 100\% & 57\% & 0\% \\
        Llama3.3-70B    & 100\% & 40\% & 0\% \\
        Gemini2.5-pro   & 100\% & 80\% & 6\% \\
        ChatGPT-o3      & 100\%  & 81\% & 3\% \\
        Claude3.5       & 100\%  & 73\% & 0\% \\
        Claude3.7       & 100\%  & 80\% & 3\% \\
        Deepseek        & 100\%  & 70\% & 0\% \\
        Deepseek(R1)    & 100\%  & 78\% & 3\% \\
        Grok            & 100\%  & 66\% & 0\% \\
        \bottomrule
    \end{tabular}
\end{table}

As shown in Table \ref{tab:landmark_model_performance}, the LLMs show a good grasp of popular landmarks. However, they perform poorly when it comes to lesser known locations. The LLMs showed a lack of knowledge when the questions became more complicated, that would require geospatial cognition, i.e. route and survey level knowledge.

\subsection*{\centering Landmark Knowledge Questions Selections}
\vspace{0.5cm}  
\label{landmark_question}
\begin{center}
\textit{The following questions are selected from a pool of 50 questions, organized into three domains: Popular Landmarks, Common Landmarks, and Geocoding. Each domain contains 5 representative questions that demonstrate typical challenges and considerations in geospatial knowledge about landmark.}
\end{center}

\vspace{1cm}  

\subsubsection*{A. Popular Landmarks}
\textit{This section contains questions related to well-known landmarks and their basic information.}\\
\textbf{Q1:} Where is the Eiffel Tower?\\
\textbf{Q2:} How big is the white house?\\
\textbf{Q3:} How tall and how heavy is the Status of Liberty?\\
\textbf{Q4:} In which year did Colosseum built?\\
\textbf{Q5:} Where is the British Museum located in London?\\

\subsubsection*{B. Common Landmarks}
\textit{This section focuses on everyday landmarks such as intersections, buildings, and natural features.}\\
\textbf{Q1:} Where is the Cathedral Church of Our Lady in Germany?\\
\textbf{Q2:} What are the names of the streets that connect Russell Square in London?\\
\textbf{Q3:} What district of Paris is Rue Saint-Jacques in?\\
\textbf{Q4:} Is Clerkenwell Road on the north or south bank of the Thames?\\
\textbf{Q5:} How many McDonald's are there in Toronto?\\

\subsubsection*{C. Geocoding of Landmarks}
\textit{This section addresses specific challenges in geocoding and coordinate verification.}\\
\textbf{Q1:} Give the coordinates of any McDonald's in Toronto.\\
\textbf{Q2:} What are the corresponding coordinates for 317 Dundas St W, Toronto, Canada?\\
\textbf{Q3:} Where is 48$^\circ$52'20.8''N, 2$^\circ$18'15.0''E? What's nearby, the more precise the better?\\
\textbf{Q4:} How far is 21 West End Ave, New York from 20 West End Ave, New York?\\
\textbf{Q5:} The straight line connecting 47$^\circ$21'13.2''N, 3$^\circ$5'10.8''E with 43$^\circ$43'14.3''N, 39$^\circ$51'18.7''E passes through which countries on the surface?\\

\begin{figure*}[t]
    \centering
    \includegraphics[width=\textwidth]{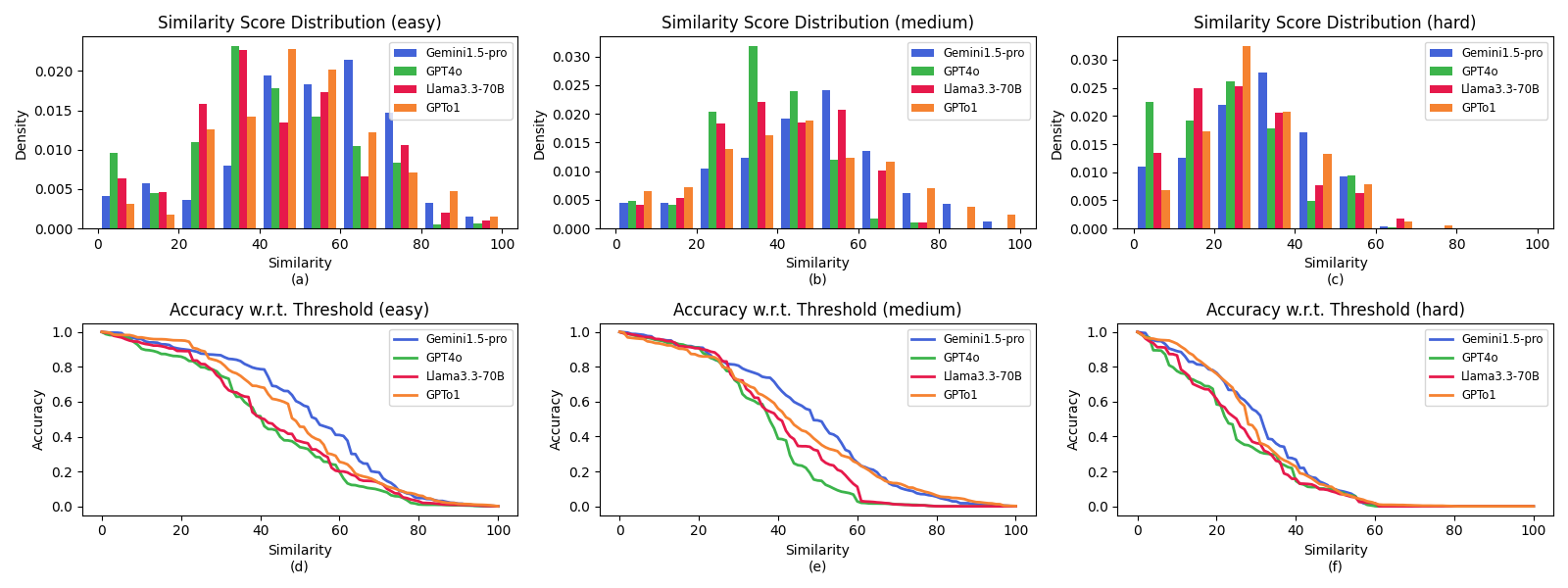}
    \caption{Performance visualization of different models across difficulty levels. The first row (a-c) shows the similarity score distributions for easy, medium, and hard datasets respectively. The second row (d-f) presents the accuracy at different similarity thresholds for each difficulty level. Each color represents a different model: Gemini2.5-pro (blue), GPT4o (green), Llama3.3-70B (red), and GPTo3 (orange).}
    \label{fig:model_performance}
\end{figure*}

\section{Prompt}

As shown in Figure \ref{fig:initial_prompt}, the guide prompt aims to help LLMs locate road networks and set the rules for route reversal. Note that we emphasized twice that the instruction in the first step must include the absolute direction, since we found in early experiments that the probability of LLMs violating the rule was much higher with just one emphasis. In addition to this, we require that LLMs use the same language style as the examples in prompt as well as no semantic inversions. However, in the final result, LLMs only adhered well to the former. It is also worth mentioning that we asked the LLMs to output some information about the surface visible along the way. In the manual evaluation, we did not find any correct examples, and in fact, LLMs tended not to answer this question.

As shown in Figure \ref{fig:instructions_prompt}, the reason for providing the latitude and longitude is to help LLMs smoothly locate the original route. Also, this further enhances the rigor of the evaluation of the geospatial aspects of LLMs. In the literature previously addressed, LLMs do possess the ability to extract some information from geographic systems. If this ability does come from an understanding of spatial relationships, and not just from textual training, then they should be able to accurately recognize the location of the routes.

\begin{figure*}[!t]  
  \centering

  \begin{subfigure}[t]{\textwidth}
    \centering
    \begin{tcolorbox}[
        colframe=black!40!gray,
        colback=black!5,
        title=Guide Prompt for Route Reversal,
        width=\textwidth,
        boxrule=0.8mm,
        sharp corners,
        coltitle=white,
        fonttitle=\bfseries
    ]
      \setlength{\parskip}{3pt}    
      \setlength{\parindent}{0pt}
      \renewcommand{\baselinestretch}{1.1}  

      Generate a road network for \texttt{[CITY NAME, COUNTRY NAME]} based on your knowledge.

      The following task involves reversing a navigation route from destination (D) back to start point (S). Follow these key requirements:

      \begin{enumerate}[leftmargin=*,itemsep=4pt]  
        \item \textbf{Start with absolute direction.} Use precise cardinal directions (North, South, East, West). Avoid ambiguous terms like ``head backward''.
        \item \textbf{No simple inversion.} Understand the route thoroughly and create logical return directions rather than merely reversing steps.
        \item \textbf{Maintain consistent format.} Use standard navigation terms (``head'', ``turn'', ``continue'', ``arrive'') as in the original directions.
        \item \textbf{Reference landmarks.} Include nearby points of interest (POI) to demonstrate geographical context.
        \item \textbf{Begin with absolute direction.} The first instruction must specify an absolute direction (non-negotiable).
      \end{enumerate}

      \textbf{Example:} \texttt{......}
    \end{tcolorbox}
    \caption{Guide prompt template for route reversal task, detailing the key requirements for generating reversed navigation instructions. The template emphasizes absolute directions, logical route understanding, format consistency, step matching, landmark referencing, and mandatory absolute direction start.}
    \label{fig:initial_prompt}
  \end{subfigure}

  \vspace{0.8em}  

  \begin{subfigure}[t]{\textwidth}
    \centering
    \begin{tcolorbox}[
        colframe=black!40!gray,
        colback=black!5,
        title=Instruction Prompt for Route Reversal,
        width=\textwidth,
        boxrule=0.8mm,
        sharp corners,
        coltitle=white,
        fonttitle=\bfseries
    ]
      \setlength{\parskip}{6pt}   
      \setlength{\parindent}{0pt}
      \renewcommand{\baselinestretch}{1.2}

      \textbf{Start Point:} 43\textdegree38'47''N, 79\textdegree26'11.5''W

      \begin{enumerate}[leftmargin=*,nosep]
        \item Head west, continue for 75.9 meters.
        \item Turn slight right, continue for 37.7 meters.
        \item Turn left, continue for 11.3 meters.
        \item Turn right, continue for 126.3 meters.
        \item Keep right, along Queen Street, continue for 91.7 meters.
        \item Keep right, continue for 146.6 meters.
        \item Turn slight left, continue for 2.3 meters.
        \item Turn right, continue for 18.8 meters.
        \item Turn right, continue for 198.7 meters.
        \item Turn left, continue for 4.8 meters.
        \item Turn right, continue for 18.7 meters.
        \item Turn right, continue for 2.1 meters.
        \item Keep left, continue for 26.4 meters.
        \item Straight ahead, then arrive at your destination.
      \end{enumerate}
    \end{tcolorbox}
    \caption{Example navigation instructions for route reversal task, showing a detailed route in Toronto with precise coordinates and step-by-step directions including distance measurements.}
    \label{fig:instructions_prompt}
  \end{subfigure}

  \caption{Prompts used in the route-reversal experiment: (a) guide prompt template with detailed requirements and (b) concrete example instructions illustrating the task.}
  \label{fig:prompts_route_reversal}
\end{figure*}


\end{document}